%% file: output.tex
\documentclass{article}
\usepackage{spconf,amsmath,graphicx}
\usepackage{amsmath,epsfig,subfigure}
\usepackage{booktabs}
\usepackage{amsfonts}
\usepackage{multirow}
\usepackage{enumitem}

\setlist[itemize]{topsep=0pt, parsep=0pt, partopsep=0pt, itemsep=0pt}

\title{AttentionLut: Attention Fusion-based Canonical Polyadic LUT for Real-time Image Enhancement}
\name{Kang Fu$^{1}$, Yicong Peng$^{1}$, Zicheng Zhang$^{1}$, Qihang Xu$^{2}$, Xiaohong Liu$^{3}$*, Jia Wang$^{1}$*, Guangtao Zhai$^{1}$}
\address{$^{1}$Institute of Image Communication and Network Engineering, Shanghai Jiao Tong University, China \\
$^{2}$ Transsion, China \\
$^{3}$ John Hopcroft Center, Shanghai Jiao Tong University, China  \\
}
\begin{document}
%
\maketitle
\begin{abstract}
Recently, many algorithms have employed image-adaptive lookup tables (LUTs) to achieve real-time image enhancement. Nonetheless, a prevailing trend among existing methods has been the employment of linear combinations of basic LUTs to formulate image-adaptive LUTs, which limits the generalization ability of these methods. To address this limitation, we propose a novel framework named AttentionLut for real-time image enhancement, which utilizes the attention mechanism to generate image-adaptive LUTs. Our proposed framework consists of three lightweight modules. We begin by employing the global image context feature module to extract image-adaptive features. Subsequently, the attention fusion module integrates the image feature with the priori attention feature obtained during training to generate image-adaptive canonical polyadic tensors. Finally, the canonical polyadic reconstruction module is deployed to reconstruct image-adaptive residual 3DLUT, which is subsequently utilized for enhancing input images. Experiments on the benchmark MIT-Adobe FiveK dataset demonstrate that the proposed method achieves better enhancement performance quantitatively and qualitatively than the state-of-the-art methods. 

\end{abstract}
\begin{keywords}
Image enhancement, Photo retouching, 3D lookup table, Attention mechanism.
\end{keywords}

\let\thefootnote\relax\footnotetext{* Corresponding Authors. E-mail:\{xiaohongliu, jiawang\}@sjtu.edu.cn. This work was supported in part by STCSM under Grant 22DZ2229005, the National Natural Science Foundation of China under Grant 62301310, the Shanghai Pujiang Program under Grant 22PJ1406800 and Transsion, China.}
\vspace{-0.2cm}
\input{Contents/Introduction}

\vspace{-0.2cm}
\input{Contents/Method}
\vspace{-0.2cm}
\input{Contents/Experiments}

\vspace{-0.2cm}
\input{Contents/Conclusion}
\bibliographystyle{IEEEbib}
\bibliography{output}

\end{document}

%% file: Contents/Introduction.tex
\section{Introduction}
\label{sec:intro}
We very likely capture low-quality photos with advanced cameras or cell phones, which is because the quality of final photos is affected by external factors such as ambient light and temperature. To procure visually pleasing images, image enhancement\cite{li2023fastllve, liu2022griddehazenet+, liu2019griddehazenet, yin2021fmsnet, huang2023transmrsr, wang2021single} technology becomes indispensable for the refinement of these low-quality photographs. The realm of image enhancement has witnessed remarkable advancements through deep learning-based methodologies, consistently achieving state-of-the-art results.
HDRNet \cite{gharbi2017deep} uses a low-resolution vision of the input image to predict a set of affine transformations in bilateral space and applies the upsampled vision of these transformations to enhance the input image. CSRNet \cite{he2020conditional} employs a conditional network to extract global features and utilizes $1 \times 1$ convolutions to enact pixel-independent transformations, which simulate global brightness adjustments and other enhancement operations.
Zeng \textit{et al.} \cite{zeng2020learning} introduced a framework that predicts an image-adaptive 3DLUT, representing pixel-independent enhancement transformations, for real-time enhancing input images. Subsequent refinements in this domain include SA-3DLUT \cite{wang2021real} and AdaInt \cite{yang2022adaint}, which respectively incorporate spatial information and image-adaptive sampling to improve the image-adaptive 3DLUT.

However, it's important to note that these methods all rely on a uniform approach to generate image-adaptive 3DLUT. This approach entails the utilization of a lightweight Convolutional Neural Network (CNN) for predicting coefficients used in linear combinations of basic 3DLUTs, ultimately yielding the image-adaptive 3DLUT through these linear combinations. This ordinary linear combination of basic LUTs limits the ability to represent complex transformations of image-adaptive 3DLUT, which leads to mediocre enhancement results.
To address these issues, we proposed a novel framework to predict a more precise image-adaptive 3DLUT. We adopt an attention mechanism-based fusion method instead of linear fusion and utilize the Canonical Polyadic (CP) decomposition to decompose 3DLUT into multi one-dimension tensors.
Our contributions are summarized as follows:
\begin{figure*}[!htbp]
    \centering
    \includegraphics[width=0.9\linewidth]{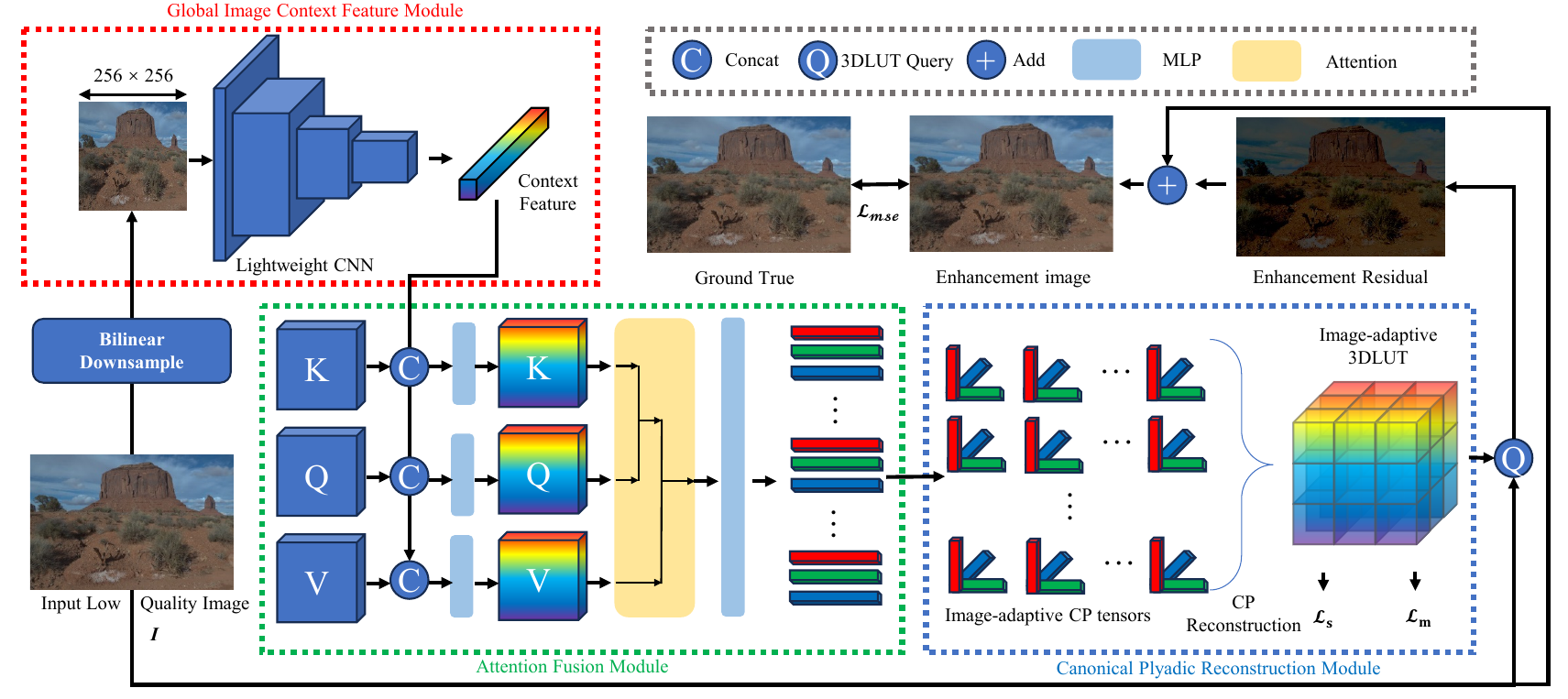}
    \caption{The framework of our proposed method. It consists of three modules: GICF module, AF module, and CPR module. The functionality of these modules are detailed in \ref{sec:GICF}, \ref{sec:AF}, and \ref{sec:CPR} respectively. The entire framework is trained in an end-to-end manner, and the loss functions are detailed in \ref{sec:loss}. }
    \label{fig:framework}
\end{figure*}

\begin{itemize}
    \item We employ an attention fusion module to fuse the global image context feature and the priori attention feature to obtain image-adaptive CP tensors.
    \item We design a canonical polyadic reconstruction module to convert the image-adaptive CP tensors to image-adaptive 3DLUT for image enhancement. This way can reduce computational complexity and parameters of the model.
    \item Experiments conducted on benchmark demonstrate that our method presents better image enhancement performance than the SOTA models.
\end{itemize}

%% file: Contents/Method.tex
\section{Methodology}
\label{sec:method}

\subsection{Preliminary: 3D Lookup Tables}
\begin{figure}[!htbp]
    \centering
    \includegraphics[width=0.9\linewidth]{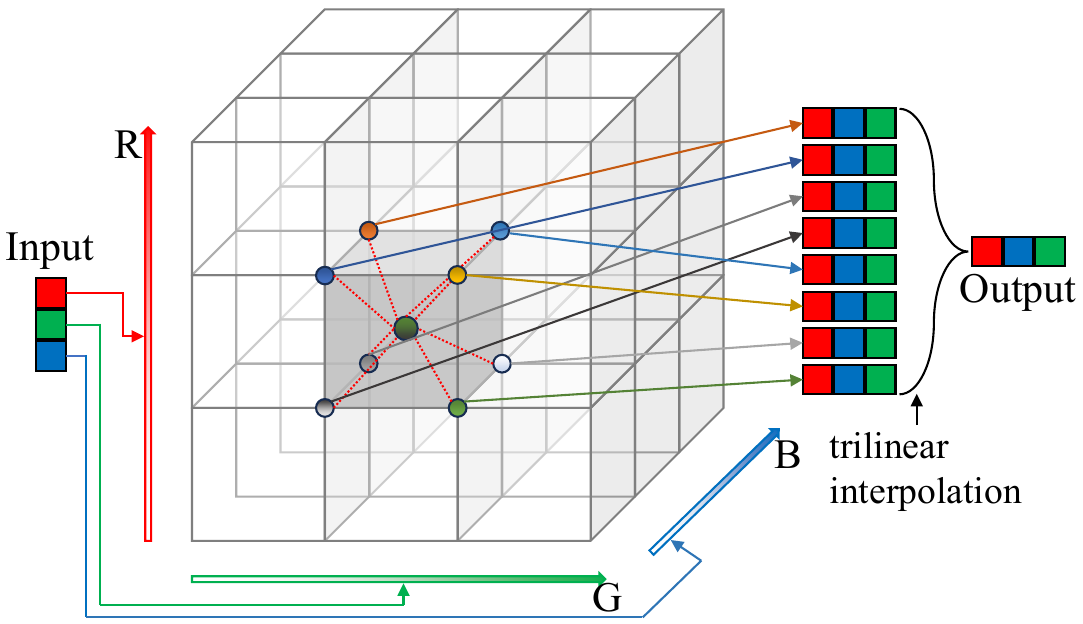}
    \caption{Schematic diagram of pixel mapping process of 3DLUT. Given a pixel, the output is obtained by trilinear interpolation through the nearest eight grid of input pixels.}
    \label{fig:3dlut}
    \vspace{-0.2cm}
\end{figure}

Traditional 3DLUT is a widely used technology for real-time color mapping or correcting. A 3DLUT is commonly represented by a 3D cubic grid $\mathbf{\Psi} \in \mathbb{R}^{3\times N \times N \times N}$ ($N$ is the number of grid in each dimension) and each grid stores the corresponding output values $\mathbf{\Psi}(x,y,z) \in \mathbb{R}^{3}$, where $x,y,z = 1, \dots , N$. Assuming an input pixel value is $(r_{i}, g_{i}, b_{i})$, the outputs $(r_{o}, g_{o}, b_{o})$ can be obtained in two steps: (1) Find the nearest eight grids with input pixel values as coordinate points. (2) Calculate the output through trilinear interpolation for these eight grids. This process is intuitively illustrated in the Fig \ref{fig:3dlut}. Given an input image $I$, we can get the enhanced image $O = \mathbf{\Psi}(I)$ by querying pixel by pixel.

\subsection{Overall Framework}

As outlined in Fig \ref{fig:framework}, Our proposed framework consists of three parts: global image context feature (GICF) module, attention fusion (AF) module, and canonical polyadic reconstruction (CPR) module. Firstly, the GICF module employs a lightweight CNN to extract the global image context feature, which is used to predict image-adaptive 3DLUT. Next, the AF module converts the priori attention feature and global image context feature to image-adaptive CP tensors. Then, the CPR Module calculates image-adaptive 3DLUT by the outer products of above CP tensors. Finally, the enhancement image is the addition of the input image and enhancement residual image which is obtained by image-adaptive residual 3DLUT $\mathbf{\Psi}_{r}$ query.

\subsection{Global Image Context Feature Module}
\label{sec:GICF}
The image enhancement algorithm using traditional 3DLUT manually selects one from a series of preset 3DLUTs for color mapping. This results in the different images being only able to use preset color mapping, which lacks generalization and flexibility. 
To solve this shortcoming, the image feature needs to be used in the calculation of image-adaptive 3DLUT. Since 3DLUT performs global color mapping on images, we chose to use a lightweight CNN and a low-resolution version of the input image to predict global image context features as in the previous work\cite{zeng2020learning}. In order to process an arbitrarily sized image in real-time, The GICF module first resizes the input low-quality image to $256 \times 256$ using bilinear interpolation and then extracts global image context feature from it through a lightweight CNN $\mathcal{G}$. This CNN consists of 5 convolutional layers with a stride of 2 and an average pooling layer. In conclusion, the GICF module can be described as follows:
\begin{equation}
    f_{I} = \mathcal{G}(\mathcal{D}(I)),
\end{equation}
where $I$ is the input image, $\mathcal{D}$ is the bilinear downsample and $f_{I} \in \mathbb{R}^{1 \times 128}$ denotes the global image context feature. 

\subsection{Attention Fusion Module}
\label{sec:AF}
It is significant to predict precise image-adaptive 3DLUT because every pixel value of the enhancement image is stored in its grid. Previous works \cite{zeng2020learning,wang2021real,yang2022adaint} are all converting image features to the linear combination coefficients of basic LUTs and representing image-adaptive 3DLUT as the linear combination of basic LUTs. This fusion approach leads to image-adaptive 3DLUT with weak generalization and limited representation of complex color mappings. Therefore, we opt to obtain better enhancement by changing the fusion method of image-adaptive 3DLUT. 

Inspired by \cite{vaswani2017attention}, we use the multi-head attention mechanism to calculate image-adaptive 3DLUT by global image context feature and priori information. To simplify calculation, we utilize \textbf{Key}, \textbf{Query}, and \textbf{Value} (abbreviated as KQV) to store prior information instead of basic LUTs. We first convert priori KQV and global image context feature to image-adaptive KQV, and then calculate image-adaptive CP tensors through image-adaptive KQV. This process can be formulated as:
\begin{equation}
    \mathcal{A}_{I} = \mathcal{P_{\mathcal{A}}}(\mathcal{A},f_{I}) \quad \mathcal{A} \in \{K,Q,V\},
\end{equation}
where $\mathcal{A}$ and $\mathcal{A}_{I}$ are respectively priori KQV and image-adaptive KQV, $P_{\mathcal{A}}$ is an ordinary linear layer, which convert $\mathcal{A}$ to $\mathcal{A}_{I}$.
\begin{equation}
    \mathcal{C} = \mathcal{Z} (softmax(\frac{Q_{I}K_{I}^\mathsf{T}}{\sqrt{d_{K_{I}}}})V_{I}),
\end{equation}
where $K_{I}$, $Q_{I}$, $V_{I}$ are respectively image-adaptive K, Q and V, $d_{K_{I}}$ is the last dimension of $K_{I}$ and $Q_{I}$, $\mathcal{Z}$ is fully-connected layer to compress intermediate feature dimensions and $\mathcal{C} \in \mathbb{R}^{9 \times X \times N}$ denotes image-adaptive CP tensors.

\subsection{Canonical Polyadic Reconstruction Module}
\label{sec:CPR}
\begin{figure}[!htbp]
    \centering
    \includegraphics[width=\linewidth]{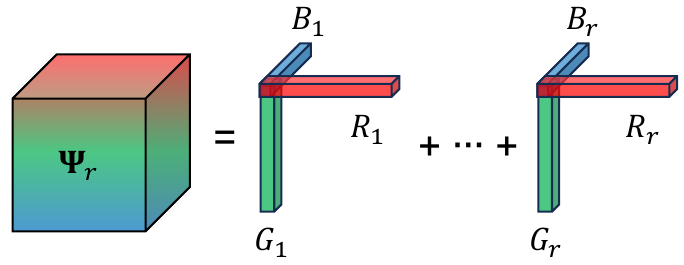}
    \caption{Illustration of canonical polyadic decomposition. It factorizes a tensor as a sum of vector outer products. }
    \label{fig:cp}
    \vspace{-0.4cm}
\end{figure}

The parameters of 3DLUT are proportional to $N^{3}$, which leads to large model size and calculation. Inspired by \cite{Chen2022ECCV}, we apply classical Canonical Polyadic decomposition \cite{carroll1970analysis} to factorize the image enhancement 3DLUT into $X$ Rank-1 tensor (shown in Fig \ref{fig:cp}). This process can be formulated as :

\begin{equation}
    \mathbf{\Psi}_{r} = \sum_{x=1}^{X} R_{x} \otimes G_{x} \otimes B_{x},
\end{equation}
where $R_{x}, G_{x}, B_{x} \in \mathbb{R}^{3 \times 1 \times N}$  are respectively the R, G, and B channels in image-adaptive CP tensors, $\otimes$ denotes the outer product.

\subsection{Loss Functions}
\label{sec:loss}
Similar to previous work \cite{zeng2020learning}, The training loss function consists of three basic losses: the MSE loss $\mathbf{\mathcal{L}_{mse}}$, the smoothness regularization loss $\mathbf{\mathcal{L}_{s}}$ and the monotonicity regularization loss $\mathbf{\mathcal{L}_{m}}$. The final loss is defined as:
\begin{equation}
    \mathcal{L} = \mathbf{\mathcal{L}_{mse}} + \lambda_{s} \mathbf{\mathcal{L}_{s}} + \lambda_{m} \mathbf{\mathcal{L}_{m}},
    \label{eq:loss}
\end{equation}
where $\lambda_{s}$ and $\lambda_{m}$ are trade-off coefficients, which are set to $1 \times 10^{-4}$ and 10. The $\mathbf{\mathcal{L}_{mse}}$ is used to optimize the performance of enhancement, The $\mathbf{\mathcal{L}_{s}}$ and $\mathbf{\mathcal{L}_{m}}$ are employed for ensuring a more natural enhancement results. Since the image-adaptive residual 3DLUT stores the pixel value of the enhancement residual, the $\mathbf{\mathcal{L}_{s}}$ and $\mathbf{\mathcal{L}_{m}}$ are calculated through the 3DLUT $\mathbf{\Psi}_{l} = \mathbf{\Psi}_{r} + \mathbf{\Psi}_{f}$, where $\mathbf{\Psi}_{f}$ is a 3DLUT where the value stored at grid $(i,j,k)$ is $(i,j,k)$.

%% file: Contents/Experiments.tex
\section{Experiments}
\label{sec:exp}
\begin{figure*}[!thb]
	\scriptsize        
	\centering
	\renewcommand{\tabcolsep}{1.0pt} %
	\begin{tabular}{ccccccc}
	
	\includegraphics[width=0.16\linewidth]{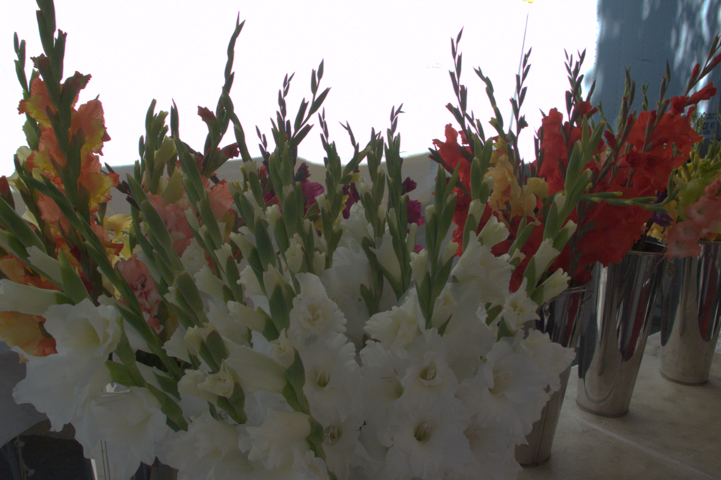}&
	\includegraphics[width=0.16\linewidth]{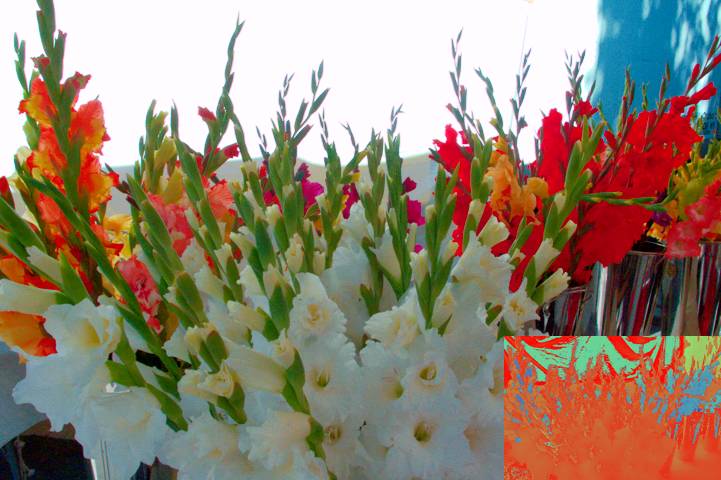}&
	\includegraphics[width=0.16\linewidth]{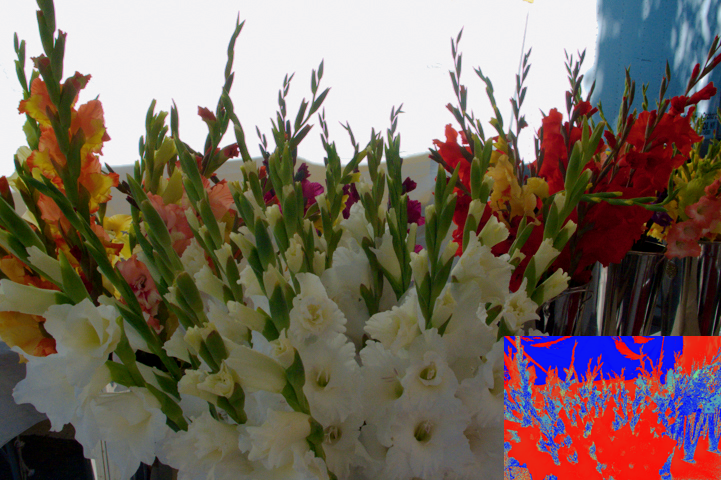}&
	\includegraphics[width=0.16\linewidth]{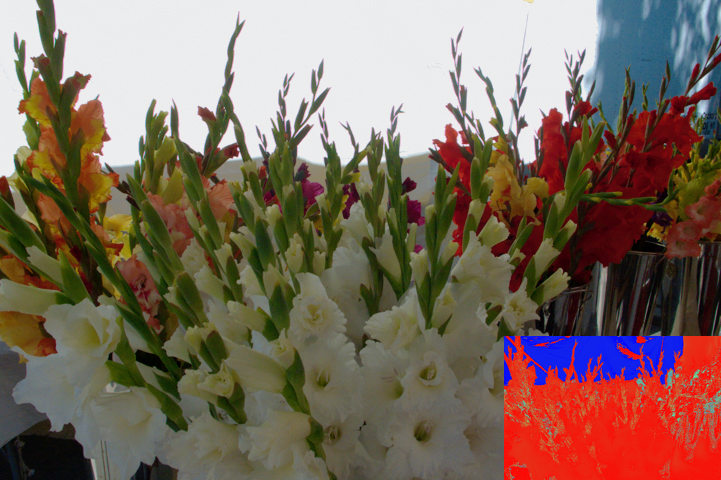}&
	\includegraphics[width=0.16\linewidth]{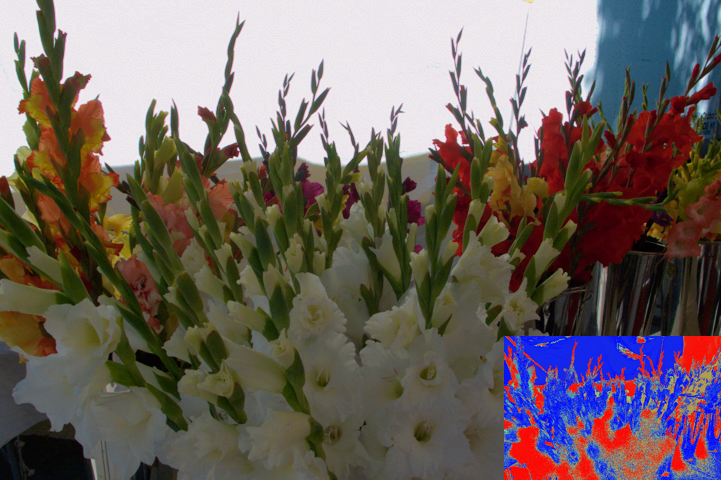}&
	\includegraphics[width=0.16\linewidth]{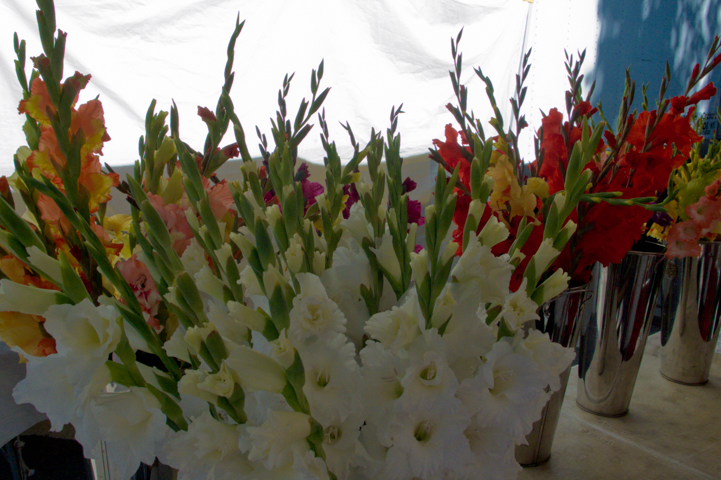}&
	\multirow{3}{*}[6.9em]{\includegraphics[width=0.0257\linewidth]{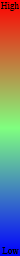}}
	\\
	
	\includegraphics[width=0.16\linewidth]{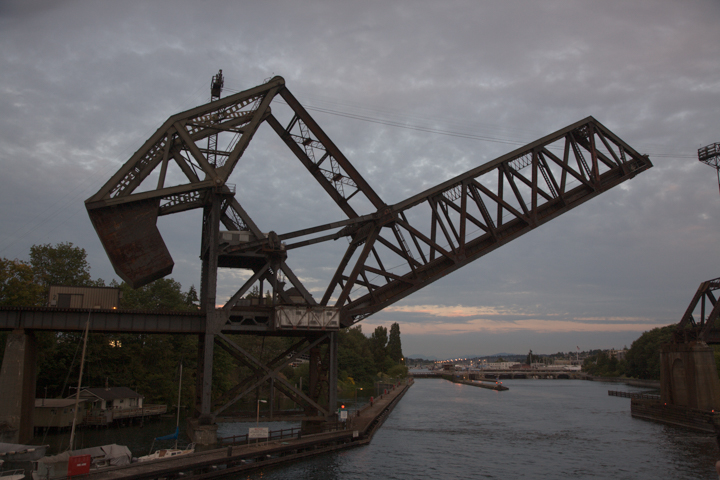}&
	\includegraphics[width=0.16\linewidth]{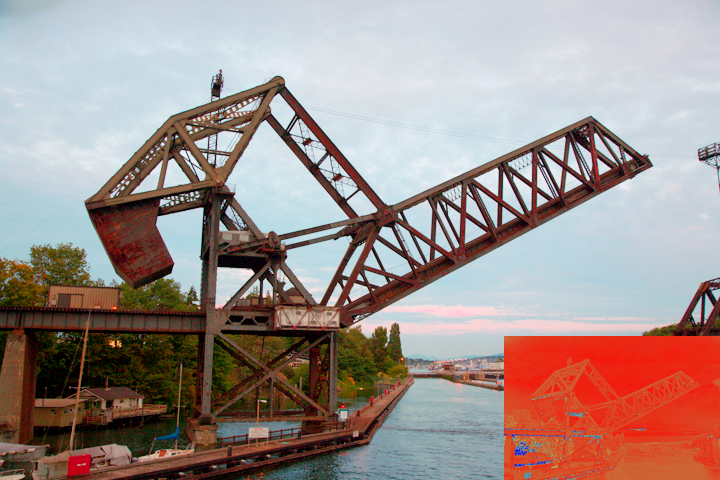}&
	\includegraphics[width=0.16\linewidth]{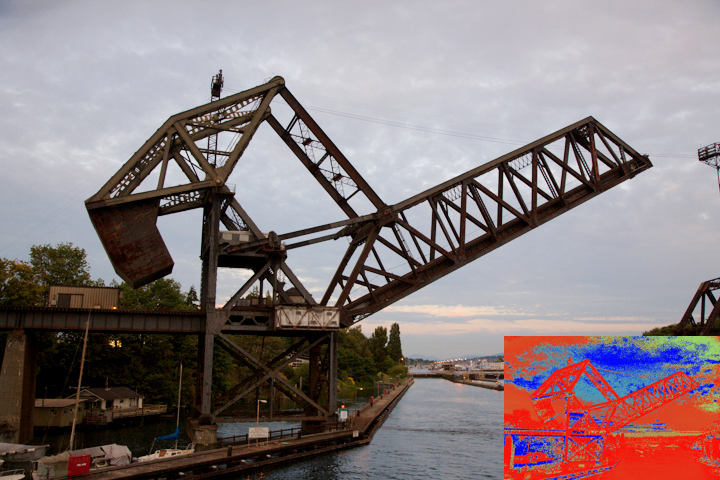}&
	\includegraphics[width=0.16\linewidth]{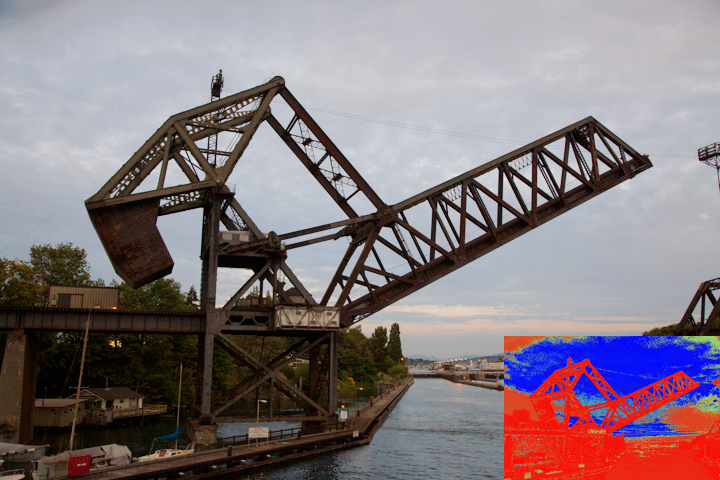}&
	\includegraphics[width=0.16\linewidth]{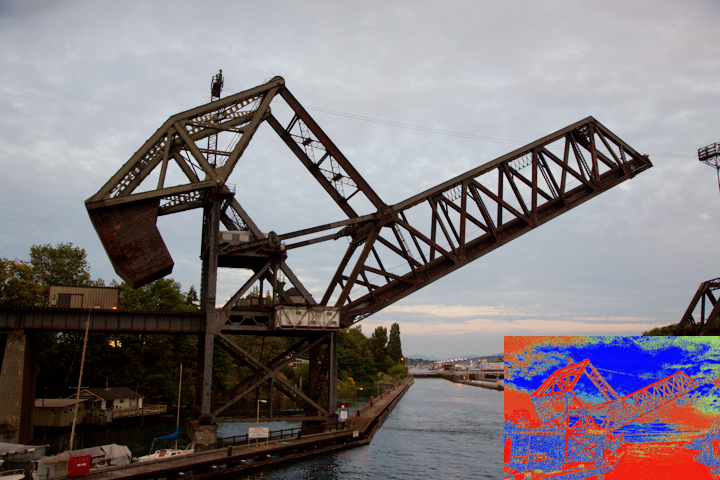}&
	\includegraphics[width=0.16\linewidth]{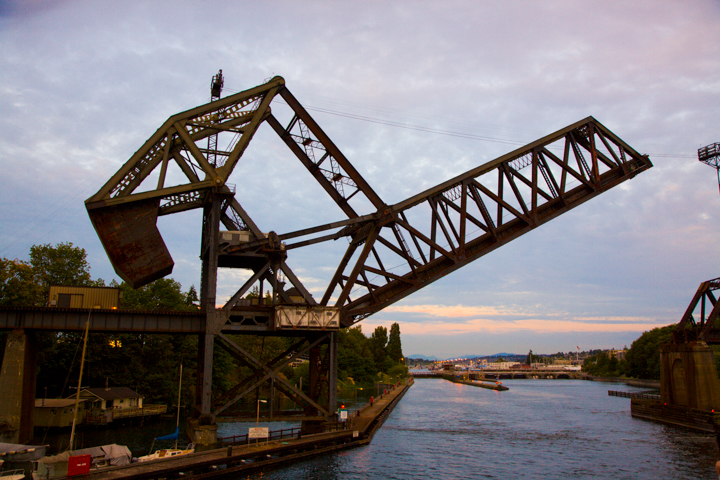}
	&
	\\
	
	\includegraphics[width=0.16\linewidth]{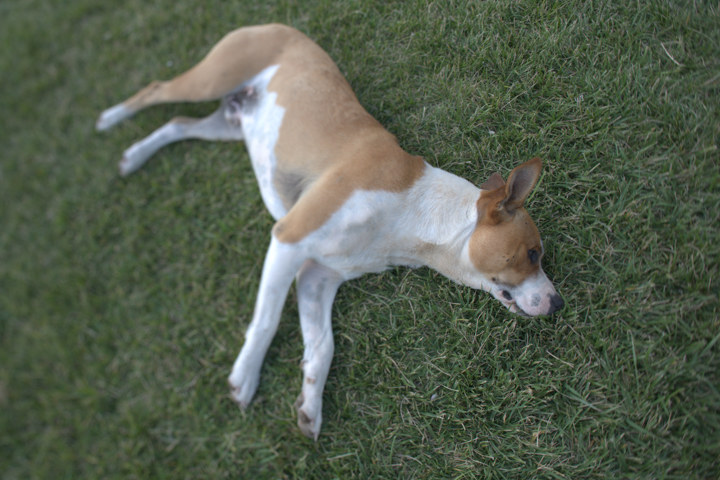}&
	\includegraphics[width=0.16\linewidth]{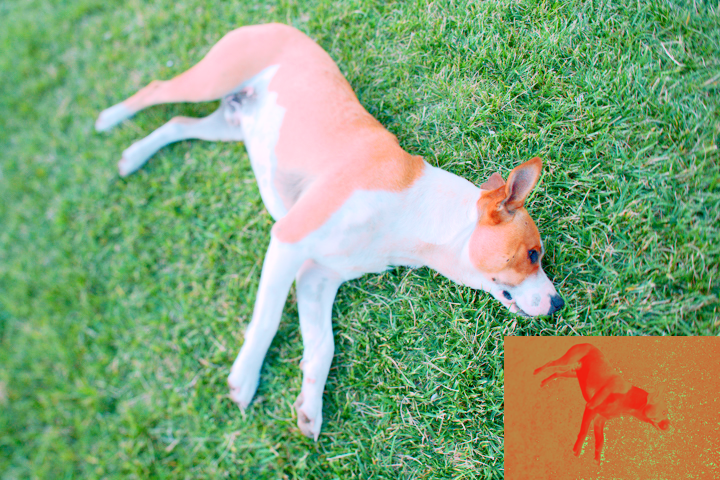}&
	\includegraphics[width=0.16\linewidth]{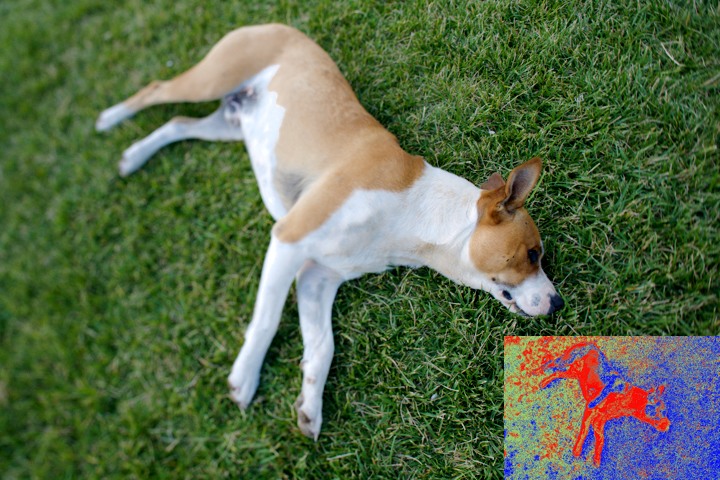}&
	\includegraphics[width=0.16\linewidth]{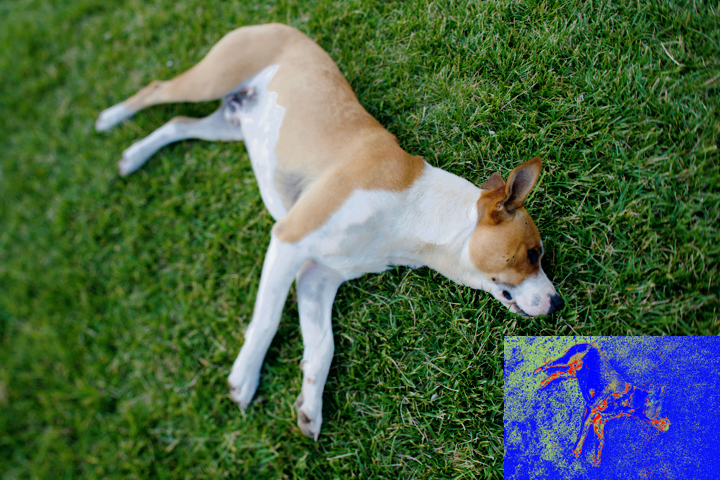}&
	\includegraphics[width=0.16\linewidth]{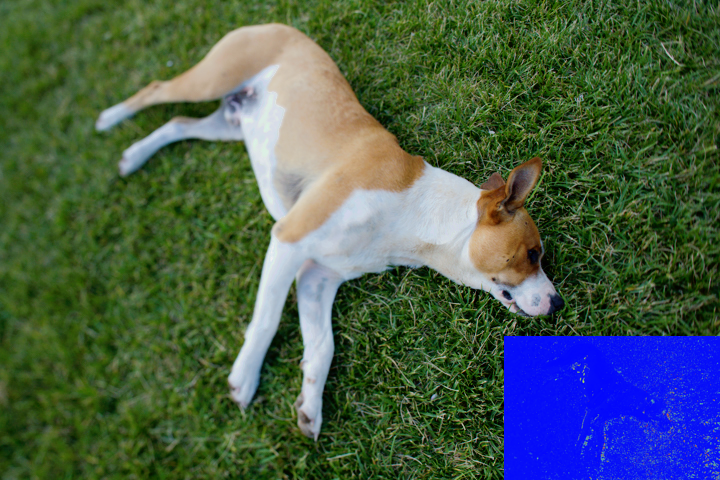}&
	\includegraphics[width=0.16\linewidth]{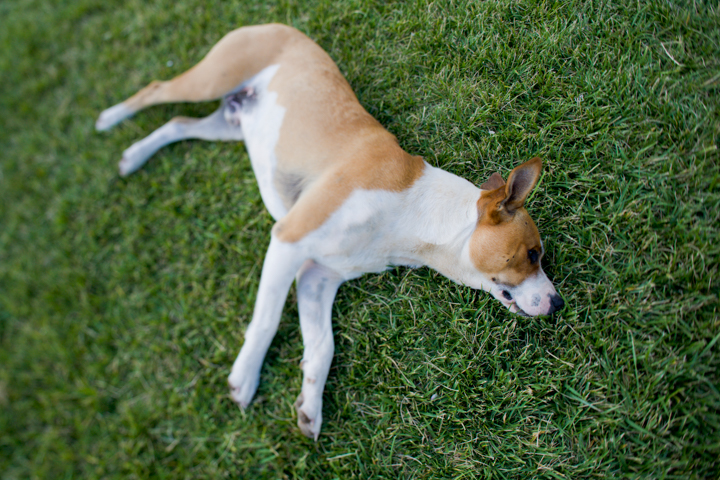}
	&
	\\
	
	(a) Input &
	(b) CSRNet &
	(c) 3DLUT &
	(d) SepLUT &
	(e) Ours &
	(f) GT &
	\\
	
    \end{tabular}
	\caption{Qualitative comparisons with corresponding error maps on the FiveK dataset for image enhancement. Blue indicates good effect and red indicates large differences. Best viewed on screen.}
	\label{fig:compare} %
    \vspace{-0.2cm}
\end{figure*}

\subsection{Experiment settings}
\textbf{Datasets} The MIT-Adobe-5K dataset\cite{bychkovsky2011learning} is a large image enhancement dataset with manually retouched ground truth, which contains 5,000 sets of images, each with a RAW image and retouched images by five human experts. Following the common practice in recent works \cite{he2020conditional, zeng2020learning}, we use the images retouched by expert C as the ground truth in subsequent experiments and divide the dataset into 4,500 training image pairs and 500 testing pairs. 

\noindent \textbf{Evaluation metrics} In order to evaluate the enhancement performance of different methods, we use PSNR, SSIM \cite{wang2004image}, $\bigtriangleup E$ as experimental evaluation metrics. $\bigtriangleup E$ is the mean L2-distance between the predicted image and ground truth in CIELAB color space.

\noindent \textbf{Implementation Details} We conduct all the experiments on an NVIDIA TITAN Xp GPU with Pytorch \cite{paszke2019pytorch} framework and use the standard Adam \cite{kingma2014adam} optimizer to minimize the loss function in Equation \ref{eq:loss}. The mini-batch size is set to 1. All our models are trained in 400 epochs with a fixed learning rate $1 \times 10^{-4}$. In order to achieve a trade-off between enhancement effect and parameter quantity, the number of CP tensors $X$ and the number of grids in each dimension $N$ are set to 15 and 33 respectively. $d_{K_{I}}$ is set to 128.


\subsection{Comparison with SOTA and Results}
We compare proposed method with a series of state-of-the-art methods including CSRNet \cite{he2020conditional}, 3DLUT \cite{zeng2020learning}, SALUT \cite{wang2021real}, DHFN \cite{zhang2023dual} and SepLUT \cite{yang2022seplut}.

\noindent \textbf{Quantitative Comparisons} As shown in Table \ref{tab:comparison}, our proposed method effectively improves the enhancement performance while ensuring the size of parameters, demonstrating the effectiveness of the AF module and CPR module.

\noindent \textbf{Qualitative Comparisons} Fig \ref{fig:compare} also shows that our method produces more visually pleasing results than other methods. The error maps indicate that our results are closer to ground truth, as the AF module makes the framework predict more precise image-adaptive 3DLUT, which represents a more precise color mapping of enhancement.

\subsection{Ablation Study}
In this section, we conduct a series of ablation experiments on the AF module and CPR Module to verify the effectiveness of these proposed modules. Specifically, we train two different models: (b) Using linear combination fusion instead of attention fusion, and (c) Using complete 3DLUT instead of CP decomposition. At the same time, we also choose two models to compare: (a) the original image-adaptive 3DLUT, and (d) the complete version of our method.
As shown in Table \ref{tab:ablation}. the AF module significantly increases the performance because it helps predict a more precise and generalized image-adaptive 3DLUT. By comparing (c) and (d), we can conclude that the CPR module effectively reduced the model size while slightly improving performance. 
\begin{table}[!htb]
\renewcommand\arraystretch{0.9}
    \centering
    \caption{Quantitative comparison with state-of-the-art methods. Best in bold.}
    \vspace{-0.2cm}
    \begin{center}
    \begin{tabular}{ccccc}
        \toprule
        Method & PSNR$\uparrow$ & SSIM$\uparrow$ & $\Delta$ E $\downarrow$ & Param.(k)$\downarrow$ \\ \hline
        CSRNet \cite{he2020conditional} & 24.70 & 0.881 & 9.69 & \textbf{37} \\
        3DLUT \cite{zeng2020learning} & 25.23 & 0.912 & 7.60 & 592 \\
        SALUT \cite{wang2021real} & 25.40 & 0.925 & \textbf{7.46} & 4,155 \\
        DHFN \cite{zhang2023dual} & 25.46 & 0.898 & 9.16 & 332 \\
        SepLUT \cite{yang2022seplut} & 25.47 & 0.925 & 7.54 & 120 \\
        Ours & \textbf{25.56} & \textbf{0.926} & 7.53 & 593\\
        \bottomrule
    \end{tabular}
    \end{center}
    \vspace{-0.4cm}
    \label{tab:comparison}
\end{table}
\vspace{-0.2cm}
\begin{table}[!htb]
\renewcommand\arraystretch{0.9}
    \centering
    \caption{Ablation study of the different modules in our framework. The best results are boldface.}
    \vspace{-0.2cm}
    \begin{center}
    \begin{tabular}{ccccc}
        \toprule
         Method &  PSNR $\uparrow$ & SSIM$\uparrow$ & $\Delta$ E $\downarrow$ & Param.(k)$\downarrow$ \\ \hline
         (a) &  25.23 & 0.912 & 7.60 & 592 \\
         (b) &  25.42 & 0.915 & 7.58 & \textbf{294} \\
         (c) &  25.47 & 0.918 & 7.57 & 3239 \\
         (d) &  \textbf{25.56} & \textbf{0.926} & \textbf{7.53} & 593 \\
         \bottomrule
    \end{tabular}
    \end{center}
    \vspace{-0.4cm}
    \label{tab:ablation}
\end{table}

%% file: Contents/Conclusion.tex
\section{Conclusion}
\label{sec:con}
\vspace{-0.1cm}
Since the existing deep learning work on image enhancement uses the linear combination of basic LUTs to represent the image-adaptive 3DLUT, which limits the expression ability of 3DLUT, this paper proposed a novel real-time image enhancement framework called AttentionLUT, which consists of three modules: GICF module, AF module, and CPR module. The GICF module is used to extract the global image context feature from the low-resolution version of the input image and the AF module converts image feature and priori attention feature to image-adaptive CP tensors. The CPR module uses CP tensors to reconstruct image-adaptive 3DLUT to enhance the input image. Extensive experiments demonstrate that our method achieves great performance in both quantitative measures and visual qualities thanks to the AF and CPR modules. 

%% file: output.bbl
\begin{thebibliography}{10}

\bibitem{li2023fastllve}
Wenhao Li, Guangyang Wu, Wenyi Wang, Peiran Ren, and Xiaohong Liu,
\newblock ``Fastllve: Real-time low-light video enhancement with
  intensity-aware look-up table,''
\newblock in {\em Proceedings of the 31st ACM International Conference on
  Multimedia}, 2023, pp. 8134--8144.

\bibitem{liu2022griddehazenet+}
Xiaohong Liu, Zhihao Shi, Zijun Wu, Jun Chen, and Guangtao Zhai,
\newblock ``Griddehazenet+: An enhanced multi-scale network with intra-task
  knowledge transfer for single image dehazing,''
\newblock {\em IEEE Transactions on Intelligent Transportation Systems}, vol.
  24, no. 1, pp. 870--884, 2022.

\bibitem{liu2019griddehazenet}
Xiaohong Liu, Yongrui Ma, Zhihao Shi, and Jun Chen,
\newblock ``Griddehazenet: Attention-based multi-scale network for image
  dehazing,''
\newblock in {\em Proceedings of the IEEE/CVF international conference on
  computer vision}, 2019, pp. 7314--7323.

\bibitem{yin2021fmsnet}
Xiangyu Yin, Xiaohong Liu, and Huan Liu,
\newblock ``Fmsnet: Underwater image restoration by learning from a synthesized
  dataset,''
\newblock in {\em Artificial Neural Networks and Machine Learning--ICANN 2021:
  30th International Conference on Artificial Neural Networks, Bratislava,
  Slovakia, September 14--17, 2021, Proceedings, Part III 30}. Springer, 2021,
  pp. 421--432.

\bibitem{huang2023transmrsr}
Shan Huang, Xiaohong Liu, Tao Tan, Menghan Hu, Xiaoer Wei, Tingli Chen, and Bin
  Sheng,
\newblock ``Transmrsr: Transformer-based self-distilled generative prior for
  brain mri super-resolution,''
\newblock {\em arXiv preprint arXiv:2306.06669}, 2023.

\bibitem{wang2021single}
Wenyi Wang, Jun Hu, Xiaohong Liu, Jiying Zhao, and Jianwen Chen,
\newblock ``Single image super resolution based on multi-scale structure and
  non-local smoothing,''
\newblock {\em EURASIP Journal on Image and Video Processing}, vol. 2021, no.
  1, pp. 16, 2021.

\bibitem{gharbi2017deep}
Micha{\"e}l Gharbi, Jiawen Chen, Jonathan~T Barron, Samuel~W Hasinoff, and
  Fr{\'e}do Durand,
\newblock ``Deep bilateral learning for real-time image enhancement,''
\newblock {\em ACM Transactions on Graphics (TOG)}, vol. 36, no. 4, pp. 1--12,
  2017.

\bibitem{he2020conditional}
Jingwen He, Yihao Liu, Yu~Qiao, and Chao Dong,
\newblock ``Conditional sequential modulation for efficient global image
  retouching,''
\newblock {\em arXiv preprint arXiv:2009.10390}, 2020.

\bibitem{zeng2020learning}
Hui Zeng, Jianrui Cai, Lida Li, Zisheng Cao, and Lei Zhang,
\newblock ``Learning image-adaptive 3d lookup tables for high performance photo
  enhancement in real-time,''
\newblock {\em IEEE Transactions on Pattern Analysis and Machine Intelligence},
  vol. 44, no. 4, pp. 2058--2073, 2020.

\bibitem{wang2021real}
Tao Wang, Yong Li, Jingyang Peng, Yipeng Ma, Xian Wang, Fenglong Song, and
  Youliang Yan,
\newblock ``Real-time image enhancer via learnable spatial-aware 3d lookup
  tables,''
\newblock in {\em Proceedings of the IEEE/CVF International Conference on
  Computer Vision}, 2021, pp. 2471--2480.

\bibitem{yang2022adaint}
Canqian Yang, Meiguang Jin, Xu~Jia, Yi~Xu, and Ying Chen,
\newblock ``Adaint: Learning adaptive intervals for 3d lookup tables on
  real-time image enhancement,''
\newblock in {\em Proceedings of the IEEE/CVF Conference on Computer Vision and
  Pattern Recognition}, 2022, pp. 17522--17531.

\bibitem{vaswani2017attention}
Ashish Vaswani, Noam Shazeer, Niki Parmar, Jakob Uszkoreit, Llion Jones,
  Aidan~N Gomez, {\L}ukasz Kaiser, and Illia Polosukhin,
\newblock ``Attention is all you need,''
\newblock {\em Advances in neural information processing systems}, vol. 30,
  2017.

\bibitem{Chen2022ECCV}
Anpei Chen, Zexiang Xu, Andreas Geiger, Jingyi Yu, and Hao Su,
\newblock ``Tensorf: Tensorial radiance fields,''
\newblock in {\em European Conference on Computer Vision (ECCV)}, 2022.

\bibitem{carroll1970analysis}
J~Douglas Carroll and Jih-Jie Chang,
\newblock ``Analysis of individual differences in multidimensional scaling via
  an n-way generalization of “eckart-young” decomposition,''
\newblock {\em Psychometrika}, vol. 35, no. 3, pp. 283--319, 1970.

\bibitem{bychkovsky2011learning}
Vladimir Bychkovsky, Sylvain Paris, Eric Chan, and Fr{\'e}do Durand,
\newblock ``Learning photographic global tonal adjustment with a database of
  input/output image pairs,''
\newblock in {\em CVPR 2011}. IEEE, 2011, pp. 97--104.

\bibitem{wang2004image}
Zhou Wang, Alan~C Bovik, Hamid~R Sheikh, and Eero~P Simoncelli,
\newblock ``Image quality assessment: from error visibility to structural
  similarity,''
\newblock {\em IEEE transactions on image processing}, vol. 13, no. 4, pp.
  600--612, 2004.

\bibitem{paszke2019pytorch}
Adam Paszke, Sam Gross, Francisco Massa, Adam Lerer, James Bradbury, Gregory
  Chanan, Trevor Killeen, Zeming Lin, Natalia Gimelshein, Luca Antiga, et~al.,
\newblock ``Pytorch: An imperative style, high-performance deep learning
  library,''
\newblock {\em Advances in neural information processing systems}, vol. 32,
  2019.

\bibitem{kingma2014adam}
Diederik~P Kingma and Jimmy Ba,
\newblock ``Adam: A method for stochastic optimization,''
\newblock {\em arXiv preprint arXiv:1412.6980}, 2014.

\bibitem{zhang2023dual}
Yuhong Zhang, Hengsheng Zhang, Li~Song, Rong Xie, and Wenjun Zhang,
\newblock ``Dual-head fusion network for image enhancement,''
\newblock in {\em ICASSP 2023-2023 IEEE International Conference on Acoustics,
  Speech and Signal Processing (ICASSP)}. IEEE, 2023, pp. 1--5.

\bibitem{yang2022seplut}
Canqian Yang, Meiguang Jin, Yi~Xu, Rui Zhang, Ying Chen, and Huaida Liu,
\newblock ``Seplut: Separable image-adaptive lookup tables for real-time image
  enhancement,''
\newblock in {\em European Conference on Computer Vision}. Springer, 2022, pp.
  201--217.

\end{thebibliography}
